\documentclass[11pt,a4paper]{article}
\usepackage[hyperref]{emnlp2018}
\usepackage{times}
\usepackage{latexsym}

\usepackage{url}

\usepackage{graphicx}
\usepackage{amsmath}
\usepackage{multirow}
\usepackage{amssymb}
\usepackage{multirow}

\aclfinalcopy 

\title{Learning Better Internal Structure of Words for Sequence Labeling}

\author{Yingwei Xin, Ethan Hart, Vibhuti Mahajan, Jean-David Ruvini \\
  eBay Research \\ 2025 Hamilton Ave, San Jose, CA 95125, USA \\
  {\tt \{yixin,ejhart,vibmahajan,jean-david.ruvini\}@ebay.com}}

\date{}

\begin{document}
\maketitle
\begin{abstract}
 Character-based neural models have recently proven very useful for many NLP tasks. However, there is a gap of sophistication between methods for learning representations of sentences and words. While most character models for learning representations of sentences are deep and complex, models for learning representations of words are shallow and simple. Also, in spite of considerable research on learning character embeddings, it is still not clear which kind of architecture is the best for capturing character-to-word representations. To address these questions, we first investigate the gaps between methods for learning word and sentence representations. We conduct detailed experiments and comparisons of different state-of-the-art convolutional models, and also investigate the advantages and disadvantages of their constituents. Furthermore, we propose IntNet, a funnel-shaped wide convolutional neural architecture with no down-sampling for learning representations of the internal structure of words by composing their characters from limited, supervised training corpora. We evaluate our proposed model on six sequence labeling datasets, including named entity recognition, part-of-speech tagging, and syntactic chunking. Our in-depth analysis shows that IntNet significantly outperforms other character embedding models and obtains new state-of-the-art performance without relying on any external knowledge or resources.
\end{abstract}

\section{Introduction}
\label{sec:intro}

Sequence labeling is the task of assigning a label or class to each element of a sequence of data, and is one of the first stages in many natural language processing (NLP) tasks. For example, named entity recognition (NER) aims to classify words in a sentence into several predefined categories of interest such as person, organization, location, etc. Part-of-speech (POS) tagging assigns a part of speech to each word in an input sentence. Syntactic chunking divides text into syntactically related, non-overlapping groups of words. Sequence labeling is a challenging problem because human annotation is very expensive and typically only a small amount of tagging data is available. 

Most traditional sequence labeling systems have been dominated by linear statistical models which heavily rely on feature engineering. As a result, carefully constructed hand-crafted features and domain-specific knowledge are widely used for solving these tasks. Unfortunately, it is costly to develop domain specific knowledge and hand-crafted features. Recently, neural networks using character-level information have been used successfully for minimizing the need of feature engineering. There are basically two threads of character-based modeling, one focuses on learning representations of sentences for semantics and syntax \cite{zhang2015character,conneau2017very}; the other focuses on learning representations of words for the purpose of eliminating hand-crafted features for word shape information \cite{lample2016neural,ma2016end}.

Two main state-of-the-art approaches of learning character representations for sequence labeling emerged from the latter thread. One is based on RNNs and uses bidirectional LSTMs or GRUs to learn forward and backward character information \cite{ling2015finding,lample2016neural,yang2017transfer}. The other approach is based on CNNs with a fixed-size window around each word to create character-level representations \cite{santos2014learning,chiu2015named,ma2016end}. However, there is a gap in the sophistication between character-based methods for learning representations of sentences compared to that of words. We found that most of the state-of-the-art character-based CNN models for words use a convolution followed by max pooling as a shallow feature extractor, which is very different from the CNN models with deep and complex architecture for sentences. In spite of considerable research on learning character embeddings, it is still not clear which kind of architecture is the best for capturing character-to-word representations.

Therefore, a number of questions remain open:

\begin{itemize}
    \item Why is there a gap between methods for learning representations of sentences and words? How can this gap be bridged?
    \item How do state-of-the-art character embedding models differ in term of performance?
    \item What kind of neural network architecture is better for learning the internal structure of a word? Deep or shallow? Narrow or wide?
\end{itemize}

To answer these questions, we first investigate the gap between learning word representations and sentence representations for convolutional architectures. The most straightforward idea is to add more convolutional layers which follows the approaches from learning representations of sentences. Interestingly, we observe the accuracy does not increase much and found that accuracy drops when we increased the depth of the network. This observation shows that learning character representations for the internal structure of words is very different than sentences, and also might explain one of the reasons there has been a gap in character-based CNN models for representing words and sentences.

In this paper, we present detailed experiments and comparisons across different state-of-the-art convolutional models from natural language processing and computer vision. We also investigate the advantages and disadvantages of some of their constituents on different convolutional architectures. Furthermore, we propose IntNet, a funnel-shaped wide convolutional neural network for learning the internal structure of words by composing their characters. Unlike previous CNN-based approaches, our funnel-shaped IntNet explores deeper and wider architecture with no down-sampling for learning character-to-word representations from limited supervised training corpora. Lastly, we combine our IntNet model with LSTM-CRF, which captures both word shape and context information, and jointly decode tags for sequence labeling.

The main contributions of this paper are the following:
\begin{itemize}

\item We conduct detailed studies on investigating the gap between learning word representations and sentence representations.
\item We provide in-depth experiments and empirical comparisons of different convolutional models and explore the advantages and disadvantages of their components for learning character-to-word representations.
\item We propose a funnel-shaped wide convolutional neural architecture with no down-sampling that focuses on learning a better internal structure of words.
\item Our proposed compositional character-to-word model combined with LSTM-CRF achieves state-of-the-art performance for various sequence labeling tasks.

\end{itemize}

This paper is organized as follows: Section~\ref{sec:related} describes multiple threads of related work. Section~\ref{sec:architecture} presents the whole architecture of the neural network. Section~\ref{sec:experiments} provides details about experimental settings and compared methods. Section~\ref{sec:results} reports model results on different benchmarks with detailed analyses and discussion.

\section{Related Work}
\label{sec:related}

There exist three threads of related work regarding the topic of this paper: (i) different convolutional architectures from different domains; (ii) character embedding models for words; (iii) sequence labeling with deep neural network.

\textbf{CNN models across domains.} Convolutional neural networks (CNNs) are very useful in extracting information from raw signals. In the area of NLP, \citet{kim2014convolutional} was the first to propose shallow CNN with word embeddings for sentence classification. \citet{zhang2015character} proposed CNN with 6 convolutional layers by directly extracting character level information for learning representations of semantic structure on sentences. Recently, \citet{conneau2017very} proposed a VDCNN architecture with 29 convolutional layers using residual connections for text classification. However, one study on randomly dropping layers for training deep residual networks, \cite{huang2016deep}, has shown that not all layers may be needed and highlighted there is some amount of redundancy in ResNet \cite{he2016deep}. Also, some research has shown promising results with wide architectures, for example, wide ResNet \cite{Zagoruyko2016wide}, Inception-ResNet \cite{szegedy2017inception} and DenseNet \cite{huang2017densely}. These models use character-level information to learn representations are for sentences, not words.

\textbf{Character embedding models.} \citet{santos2014learning} proposed a CNN model to learn character representations of words to replace hand-crafted features for part-of-speech tagging. \citet{ling2015finding} proposed a bidirectional LSTM over characters to use as input for learning character-to-word representations. \citet{chiu2015named} proposed a bidirectional LSTM-CNN with lexicons for named entity recognition by applying the CNN-based character embedding model from \citet{santos2014learning}. \citet{plank2016multilingual} proposed a bi-LSTM model with auxiliary loss for multilingual part-of-speech tagging by following the LSTM-based character embedding model from \citet{ling2015finding}. \citet{cotterell2017crosslingual} proposed a character-level transfer learning model for neural morphological tagging.

\textbf{Sequence labeling.} \citet{collobert2011natural} first proposed a method based on CNN-CRF that learns important features from words and requires few hand-crafted features. \citet{huang2015bidirectional} proposed a bidirectional LSTM-CRF model by using word embeddings and hand-crafted features for sequence tagging. \citet{lample2016neural} applied the LSTM-based character embedding model from \citet{ling2015finding} with bidirectional LSTM-CRF and obtained best results on NER for Spanish, Dutch, and German. \citet{ma2016end} applied the CNN-based character embedding model from \citet{chiu2015named}, but without using any data preprocessing or external knowledge and achieved the best result on NER for English and part-of-speech tagging. Also, there have been some joint models which use additional knowledge, like transfer learning \cite{yang2017transfer}, pre-trained language models \cite{peters2017semisupervised}, language model joint training \cite{rei2017semisupervised}, and multi-task learning \cite{liu2017empower}. Without any additional supervision or extra resources, LSTM-CRF \cite{lample2016neural} and LSTM-CNN-CRF \cite{ma2016end} are current state-of-the-art methods. To test the effectiveness of our proposed model, we use these two models as our baselines in the latter sections.

\section{Neural Network Architecture}
\label{sec:architecture}

\subsection{IntNet}

\textbf{Character embeddings.} The first step is to initialize the character embeddings for each word $w$ in the input sequence. We define the finite set of characters $V^{char}$. This vocabulary contains all the variations of the raw text, including uppercase and lowercase letters, numbers, punctuation marks, and symbols. Unlike some character-based approaches, we do not use any character-level prepossessing which enables our model to learn and capture regularities from prefixes to suffixes to construct character-to-word representations. The input word $w$ is decomposed into a sequence of characters \{$c_1, . . . , c_n$\}, where $n$ is the length of $w$. Character embeddings are encoded by column vectors in the embedding matrix $W^{char} \in \mathbb{R}^{d^{char} \times \mid V^{char} \mid}$, where $d^{char}$ is the number of parameters for each character in $V^{char}$. Given a character $c_i$, its embedding $r^{char}_i$ is obtained by the matrix-vector product:

\begin{equation}
r^{char}_i = W^{char} v^{char}_i,
\label{E:char_emb}
\end{equation}
where $v^{char}_i$ is defined as a one-hot vector for $c_i$. We randomly initialize a look-up table with values drawn from a uniform distribution with range $\lbrack -\sqrt{\frac{3}{d^{char}}}, +\sqrt{\frac{3}{d^{char}}} \rbrack$, where $d^{char}$ is empirically chosen by users. The character set includes all unique characters and the special tokens {\tt PADDING} and {\tt UNKNOWN}. We do not perform any character-level preprocessing, including case normalization, digit replacement (e.g. replacing all sequences of digits 0-9 with a single ``0''), nor do we use any capitalization features (e.g. {\tt allCaps}, {\tt upperInitial}, {\tt lowercase}, {\tt mixedCaps}, {\tt noinfo}).

\textbf{Convolutional blocks.} The input for the IntNet is the sequence of character embeddings \{$r^{char}_1, . . . , r^{char}_n$\}. First is the initial convolutional layer, which is a temporal convolutional module that computes 1-D convolutions. Let $\textbf{x}_{i} \in \mathbb{R}^{d^{char} \times r^{char}}$ be the concatenation of the character embeddings for each $w$. The initial convolutional layer applies a matrix-vector operation to each successive window of size $k^{char}$. An input $k$-grams $\textbf{x}_{i:i+k-1}$ is transformed through a convolution filter $\textbf{w}_{c}$:

\begin{equation}
\textbf{c}_{i} = f(\textbf{w}_c \cdot \textbf{x}_{i:i+k-1} + b_c),
\label{E:conv}
\end{equation}
where $\textbf{c}_i$ is the feature map of 1-D convolution,  $f$ is the non-linear ReLU function, and $b_c$ is a bias term. Equation \ref{E:conv} produces $m$ filters with different kernel sizes. The filters are computed with different kernels by the initial convolutional layers are concatenated: 

\begin{equation}
\textbf{g}_{0} = [c^{k_1}_1\dots c^{k_1}_m; c^{k_2}_1\dots c^{k_2}_m; c^{k_h}_1\dotsc^{k_h}_m],
\label{E:concat-feature}
\end{equation}
where $h$ is the number of kernels, $\textbf{g}_{0}$ is the output for the initial convolutional layer which feeds into the next convolutional block. 

We define $\mathcal{F}(\cdot)$ as a function of several consecutive operations within a convolutional block. Firstly, a N$\times$1 convolution transforms the input. The output size is ${4}\times{m}\times{h}$ feature maps, like a bottleneck layer. The next step consists of multiple 1-D convolutions with kernels of different sizes. Lastly, we concatenate all the feature maps from kernels of different size. In each convolution, we use a batch normalization, followed by a ReLU activation and N$\times$k temporal convolution.

\begin{figure}[!htb]
  \centering
  \includegraphics[scale=0.85]{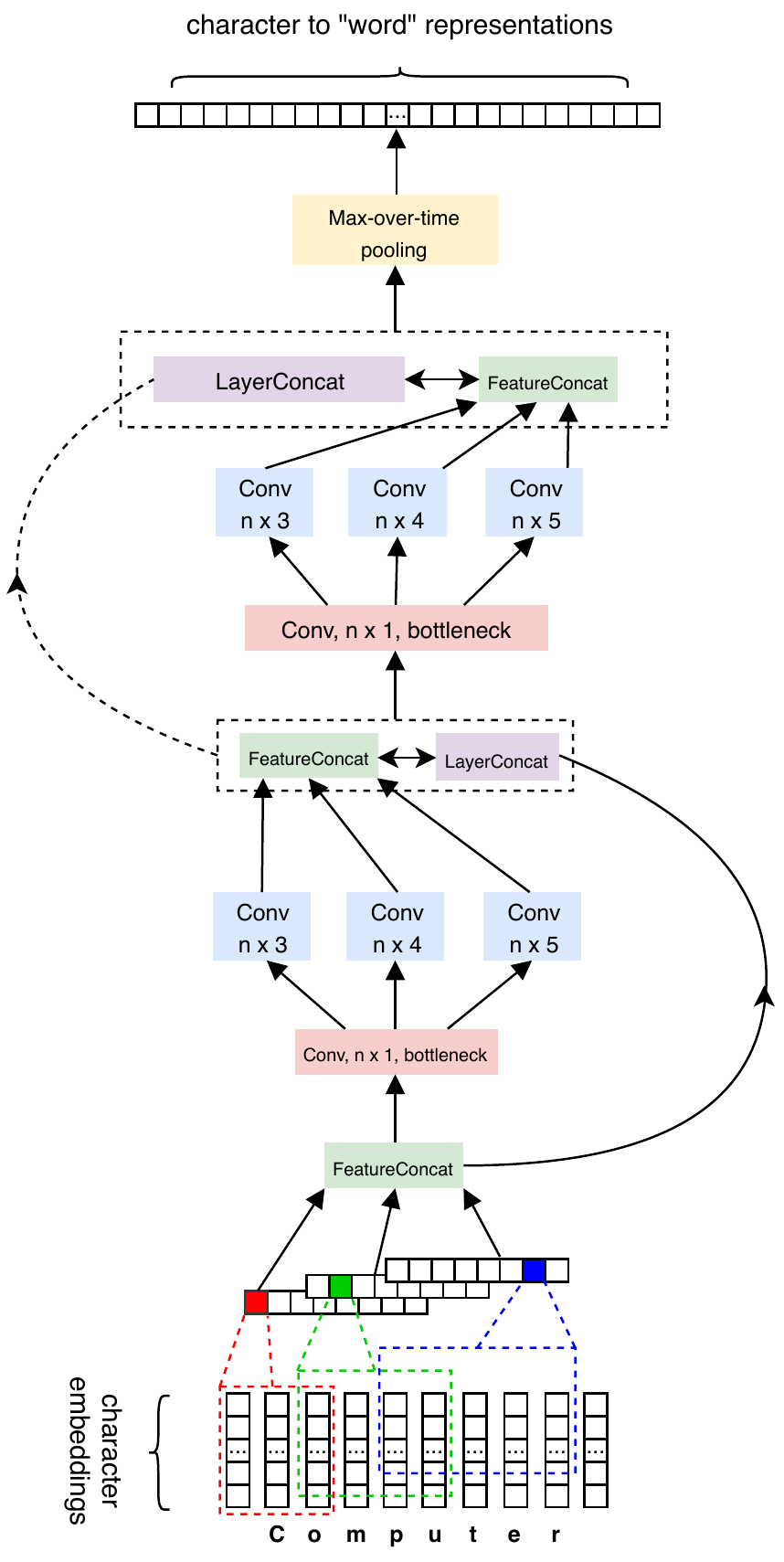}
  \caption{The main architecture of IntNet.}
  \label{fig-DCC2W}
\end{figure}

\textbf{Funnel-shaped wide architecture.} The network comprises of $L$ convolutional layers, which implies $(\frac{L-1}{2})$ convolutional blocks. We use direct connections from every other layer to all subsequent layers, inspired by dense connection. Therefore, the $l^{th}$ layer has access to the feature maps of all the alternate layers:

\begin{equation}
\textbf{g}_{l} = \mathcal{F}_{l}([\textbf{g}_0, \textbf{g}_2, \dots, \textbf{g}_{l-2}]).
\label{E:concat-layer}
\end{equation}

Equation \ref{E:concat-layer} ensures maximum information flow between blocks in the network. Compared to residual connection $\mathcal{F}_{l}(\textbf{g}_{l-1})+\textbf{g}_{l-1}$, it can be viewed as an extreme case of residual connection and makes feature reuse possible. Unlike DenseNet and ResNet, we concatenate feature maps by different kernels in every other convolutional layers, which captures different levels of features and makes our wide architecture possible, inspired by Inception. Different levels of concatenation can help IntNet to learn different patterns of word shape information. We compare our architecture to residual connection and dense connection for learning character-to-word representations in Section \ref{sec:results}. 

\textbf{Without down-sampling.} Compared to other CNN models like ResNet and DenseNet, our model does not contain any halve down-sampling layer or average pooling to reduce resolution. We did not find these operations to be helpful and, in some cases, found them to be detrimental to performance. These operations are useful for sentences and images, but might break the internal structure of words, like the sequential patterns for prefixes and suffixes. 

\textbf{Character-to-word representations.} In the last layer, we use a max-over-time pooling operation:

\begin{equation}
\hat{\textbf{c}_i} = max(\textbf{c}_i),
\label{E:maxpooling}
\end{equation}
which takes the maximum value corresponding to a particular filter. The idea is to capture the most important feature with the highest value for each feature map. Finally, we concatenate all of salient features together as a representation for this word: 

\begin{equation}
\textbf{z} = [\hat{\textbf{c}_0}, \hat{\textbf{c}_1}, \dots \hat{\textbf{c}_u}],
\label{E:c2w}
\end{equation}
where $u$ is the number of salient features which is equal to the total number of output feature maps in the last layer. If each function $\mathcal{F}_{l}$ produces $p$ feature maps, we obtain $(p_0 + p \times \frac{L-1}{2})$ representations, where $p_0$ is the number of output feature maps in the initial convolution layer.

\subsection{Bi-directional RNN}

Given the character-to-word representations are computed by IntNet in Equation \ref{E:c2w}, we denote the input vector ($\textbf{z}_{1}$, $\textbf{z}_{2}$, . . . , $\textbf{z}_{n}$) for a sentence. LSTM \cite{hochreiter1997long} returns the sequence ($\textbf{h}_{1}$, $\textbf{h}_{2}$, . . . , $\textbf{h}_{n}$) that represents the sequential information at every step. We use the following implementation:

\begin{align*}
& \textbf{i}_{t} = \sigma(\textbf{W}_{zi}\textbf{z}_{t} + \textbf{W}_{hi}\textbf{h}_{t-1} + \textbf{W}_{ci}\textbf{c}_{t-1} + \textbf{b}_{i}) \\
& \textbf{f}_{t} = \sigma(\textbf{W}_{zf}\textbf{z}_{t} + \textbf{W}_{hf}\textbf{h}_{t-1} + \textbf{W}_{cf}\textbf{c}_{t-1} + \textbf{b}_{f}) \\
& \widetilde{\textbf{c}_{t}} = tanh(\textbf{W}_{zc}\textbf{z}_{t} + \textbf{W}_{hc}\textbf{h}_{t-1} + \textbf{b}_{c}) \\
& \textbf{c}_{t} = \textbf{f}_{t} \odot \textbf{c}_{t-1} + \textbf{i}_{t} \odot \widetilde{\textbf{c}_{t}} \\
& \textbf{o}_{t} = \sigma(\textbf{W}_{zo}\textbf{z}_{t} + \textbf{W}_{ho}\textbf{h}_{t-1} + \textbf{W}_{co}\textbf{c}_{t} + \textbf{b}_{o}) \\
& \textbf{h}_{t} = \textbf{o}_{t} \odot tanh(\textbf{c}_{t}), \\
\label{E:lstm}
\end{align*}
where $\sigma$ is the element-wise sigmoid function and $\odot$ is the element-wise product. $\textbf{z}_{t}$ is the input vector at time $t$ and $\textbf{i}_{t}$, $\textbf{f}_{t}$, $\textbf{o}_{t}$, $\textbf{c}_{t}$ are the input gate, forget gate, output gate, and cell vectors, all of which are the same size as the hidden vector $\textbf{h}_{t}$. $\textbf{W}_{zi}$, $\textbf{W}_{zf}$, $\textbf{W}_{zo}$, $\textbf{W}_{zc}$ denote the weight matrices of different gates for input $\textbf{z}_{t}$; $\textbf{W}_{hi}$, $\textbf{W}_{hf}$, $\textbf{W}_{ho}$, $\textbf{W}_{hc}$ are the weight matrices for hidden state $\textbf{h}_{t}$, and $\textbf{b}_{i}$, $\textbf{b}_{f}$, $\textbf{b}_{o}$, $\textbf{b}_{c}$ denote the bias vectors. Forward LSTM and backward LSTM compute the representations of $\overrightarrow{\textbf{h}_{t}}$ and $\overleftarrow{\textbf{h}_{t}}$ for left and right context of the sentence, respectively. We concatenate two hidden states to form the output of bi-directional LSTM  [$\overrightarrow{\textbf{h}_{t}},\overleftarrow{\textbf{h}_{t}}$] for capturing context information from both sides. 

\subsection{Scoring Function}
Instead of predicting each label independently, we consider the correlations between labels in neighborhoods and jointly decode the best chain of labels for a given input sentence by leveraging a conditional random field \cite{lafferty2001conditional}. Formally, the sequence of labels is defined as:

\begin{equation}
\textbf{y} = ({y}_{1}, {y}_{2}, . . . , {y}_{T}).
\label{E:crf-label}
\end{equation}

To define the scoring function $f$($\textbf{h}$, $\textbf{y}$) for each position $t$, we multiply the hidden state $\textbf{h}^w_t$ with a parameter vector $\textbf{w}_{y_{t}}$ that is indexed by the tag ${y}_{t}$ to obtain the matrix of scores output by the bidirectional LSTM network. Therefore, the function $f$ can be written as:

\begin{equation}
f(\textbf{h}, \textbf{y}) = \sum\limits_{t=1}^{T} \textbf{w}_{y_{t}} \textbf{h}^w_t + \sum\limits_{t=1}^{T} \textbf A_{y_{t-1},y_t}.
\label{E:crf-score}
\end{equation}

In Equation \ref{E:crf-score}, $\textbf{A}$ is a matrix of transition scores, $\textbf A_{i,j}$ represents the score of a transition from the tag $i$ to tag $j$, $y_{1}$ is the start tag of a sentence. Let $\mathcal{Y}$($\textbf{h}$) denote the set of possible label sequences for $\textbf{h}$. A probabilistic model for a sequence defines a family of conditional probabilities $p(\textbf{y}|\textbf{h})$ over all possible label sequences $\textbf{y}$ given $\textbf{h}$ with the following form:

\begin{equation}
p(\textbf{y}|\textbf{h}) = \frac{e^{f(\textbf h, \textbf y)}} {\sum_{y'\in\mathcal{Y}(\textbf h)} e^{f(\textbf{h}, \textbf y')}}.
\label{E:crf-prob}
\end{equation}

\subsection{Objective Function and Inference}
For end-to-end network training, we use maximum conditional likelihood estimation to maximize the log probability of the correct tag sequence:

\begin{equation*}
log(p(\textbf{y}|\textbf{h})) = f(\textbf{h}, \textbf{y}) - log\left( \sum\limits_{y'\in\mathcal{Y}(\textbf h)} e^{f(\textbf{h}, \textbf y')} \right).
\label{E:crf-loss}
\end{equation*}

While decoding, we predict the label sequence that obtains the highest score given by:

\begin{equation}
\textbf{y}^{*} = \arg\max_{y'\in\mathcal{Y}(\textbf h)} f(\textbf{h}, \textbf y').
\end{equation}

The objective function and its gradients can be efficiently computed by dynamic programming; for inference, we use the Viterbi algorithm to find the best tag path which maximizes the score.

\section{Experiments}
\label{sec:experiments}

\begin{table*}[htb]
\centering
\scalebox{0.89}{
\begin{tabular}{l|c|c|c|c|c|c}
Model & Spanish NER & Dutch NER & English NER & German NER & Chunking & PTB POS \\ \hline                                
                                 
Baseline          & 70.73$\pm$0.42  & 63.49$\pm$0.42  & 77.51$\pm$0.39  & 54.07$\pm$0.42  & 91.97$\pm$0.21  & 95.76$\pm$0.13 \\
+ char-LSTM       & 79.93$\pm$0.43  & 77.16$\pm$0.47  & 83.98$\pm$0.46  & 64.29$\pm$0.47  & 93.31$\pm$0.23  & 97.14$\pm$0.11 \\
+ char-CNN        & 79.78$\pm$0.41  & 76.43$\pm$0.48  & 83.85$\pm$0.38  & 63.53$\pm$0.41  & 92.67$\pm$0.24  & 97.02$\pm$0.12  \\ \hline
+ char-CNN-5      & 79.63$\pm$0.38  & \textbf{76.92}$\pm$0.42  & 83.60$\pm$0.39  & \textbf{64.26}$\pm$0.42 & \textbf{93.11}$\pm$0.26 & \textbf{97.15}$\pm$0.12 \\
+ char-CNN-9      & 79.25$\pm$0.56  & 74.82$\pm$0.46  & 83.31$\pm$0.47  & 63.97$\pm$0.46 & \textbf{92.92}$\pm$0.27 & \textbf{97.13}$\pm$0.13 \\
+ char-ResNet-9   & 74.34$\pm$0.45  & \textbf{76.54}$\pm$0.39  & \textbf{83.91}$\pm$0.42  & \textbf{66.15}$\pm$0.44 & \textbf{93.85}$\pm$0.24     & 96.99$\pm$0.15 \\ 
+ char-DenseNet-9 & 78.25$\pm$0.52  & \textbf{76.71}$\pm$0.53  & \textbf{84.16}$\pm$0.41  & \textbf{67.54}$\pm$0.46 & \textbf{93.82}$\pm$0.25     & \textbf{97.13}$\pm$0.11 \\ \hline
+ char-IntNet-9   & 78.53$\pm$0.44  & \textbf{76.93}$\pm$0.47  & 83.83$\pm$0.44  & \textbf{70.11}$\pm$0.41 & \textbf{93.94}$\pm$0.26 & \textbf{97.19}$\pm$0.12 \\
+ char-IntNet-5   & \textbf{80.44}$\pm$0.43 & \textbf{78.06}$\pm$0.45 & \textbf{85.34}$\pm$0.39  & \textbf{69.48}$\pm$0.42  &  \textbf{94.27}$\pm$0.23 & \textbf{97.23}$\pm$0.11 \\ 
\end{tabular}
}
\caption{F1 score of different character-to-word models.}
\label{tab-compared-models}
\end{table*}

\subsection{Datasets} We performed experiments on six standard datasets for sequence labeling tasks, i.e. named entity recognition, part-of-speech tagging, and syntactic chunking. To test the effectiveness of our proposed model, we do not use language-specific resources (such as gazetteers), external knowledge (such as transfer learning, joint training), hand-crafted features, or any character preprocessing, we do not replace any rare words into {\tt UNKNOWN}.

\textbf{Named entity recognition.} CoNLL-2002 and CoNLL2003 datasets \cite{tksintro2002conll,tjongkimsang2003conll} contain named entity labels for Spanish, Dutch, English and German as separate datasets. These four datasets contain different types of named entities: locations, persons,
organizations, and miscellaneous entities. Unlike some approaches, we do not combine the validation set with the training set. Although POS tags were made available for these datasets, we do not leverage those as additional information which sets our approach apart from that of transfer learning. 

\textbf{Part-of-speech tagging.} The Wall Street Journal (WSJ) portion of Penn Treebank (PTB) \cite{Marcus93buildinga} contains 25 sections and categorizes each word into one out of 45 POS tags. We adopt the standard split and use sections 0-18 as training data, sections 19-21 as development data, and sections 22-24 as test data.

\textbf{Syntactic chunking.} The CoNLL 2000 chunking task \cite{tjong2000introduction} uses sections 15-18 from the Wall Street Journal corpus for training and section 20 for testing. It defines 11 syntactic chunk types (e.g., NP, VP, ADJP), we adopt the standard split and sample 1000 sentences from the training set as the development set.

\subsection{Training Settings}

\textbf{Initialization.} The size of the dimensions of character embeddings is 32 which are randomly initialized using a uniform distribution. We adopt the same initialization method for randomly initialized word embeddings that are updated during training. For IntNet, the filter size of the initial convolution is 32 and that of other convolutions is 16. We have used filters of size $[3, 4, 5]$ for all the kernels. The number of convolutional layers are 5 and 9 for IntNet-5 and IntNet-9, respectively, and we have adopted the same weight initialization as that of ResNet. We use pre-trained word embeddings for initialization, GloVe \cite{pennington2014glove} 100-dimension word embeddings for English, and fastText \cite{bojanowski2017enriching} 300-dimension word embeddings for Spanish, Dutch, and German. The state size of the bi-directional LSTMs is set to 256. We adopt standard BIOES tagging scheme for NER and Chunking.

\textbf{Optimization.} We employ mini-batch stochastic gradient descent with momentum. The batch size, momentum and learning rate are set to 10, 0.9 and $\eta_{t} = \frac{\eta_{0}}{1+\rho t}$, where $\eta_{0}$ is the initial learning rate 0.01 and $\rho = 0.05$ is the decay ratio, the value of gradient clipping is 5. Dropout is applied on the input of IntNet, LSTMs, and CRF, and its ratio 0.5 is fixed, but with no dropout inside of IntNet.

\begin{table*}[htb]
\centering
\scalebox{0.67}{
\begin{tabular}{l|c|c|c|c|c|c}
\hline
Model & Spanish   & Dutch  & English & German & Chunking & POS \\ \hline                                
                                 
Conv-CRF+Lexicon \cite{collobert2011natural}   & -     & -     & 89.59 & -     & 94.32 & 97.29 \\
LSTM-CRF+Lexicon \cite{huang2015bidirectional} & -     & -     & 90.10 & -     & 94.46 & 97.43 \\
LSTM-CRF+Lexicon+char-CNN \cite{chiu2015named} & -     & -     & 90.77 & -     & -     & - \\
LSTM-Softmax+char-LSTM \cite{ling2015finding}  & -     & -     & -     & -     & -     & 97.55 \\
LSTM-CRF+char-LSTM \cite{lample2016neural}     & 85.75 & 81.74 & 90.94 & 78.76 & -     & - \\
LSTM-CRF+char-CNN \cite{ma2016end}             & -     & -     & 91.21 & -     & -     & 97.55 \\ 
GRM-CRF+char-GRU \cite{yang2017transfer}       & 84.69 & 85.00 & 91.20 & -     & 94.66 & 97.55 \\ \hline
LSTM-CRF                & 80.33$\pm$0.37 & 79.87$\pm$0.28  & 88.41$\pm$0.22  & 73.42$\pm$0.39  & 94.29$\pm$0.11  & 96.63$\pm$0.08 \\ 
LSTM-CRF+char-LSTM      & 86.12$\pm$0.34 & 87.13$\pm$0.25  & 91.13$\pm$0.15  & 78.31$\pm$0.35  & 94.97$\pm$0.09  & 97.49$\pm$0.04 \\
LSTM-CRF+char-CNN       & 85.91$\pm$0.38 & 86.69$\pm$0.22  & 91.11$\pm$0.14  & 78.15$\pm$0.31  & 94.91$\pm$0.08 & 97.45$\pm$0.03 \\ \hline
LSTM-CRF+char-IntNet-9  & 85.71$\pm$0.39 & \textbf{87.38}$\pm$0.27 & \textbf{91.39}$\pm$0.16   & \textbf{79.43}$\pm$0.33  & \textbf{95.08}$\pm$0.07 & 97.51$\pm$0.04 \\ 
LSTM-CRF+char-IntNet-5  & \textbf{86.68}$\pm$0.35 & \textbf{87.81}$\pm$0.24 & \textbf{91.64}$\pm$0.17  & 78.58$\pm$0.32 & \textbf{95.29}$\pm$0.08 & \textbf{97.58}$\pm$0.02 \\ \hline
\end{tabular}
}
\caption{F1 score of our proposed models in comparison with state-of-the-art results.}
\label{tab-state-of-the-art}
\end{table*}

\begin{figure*}[htb]
  \centering
  \includegraphics[scale=0.222]{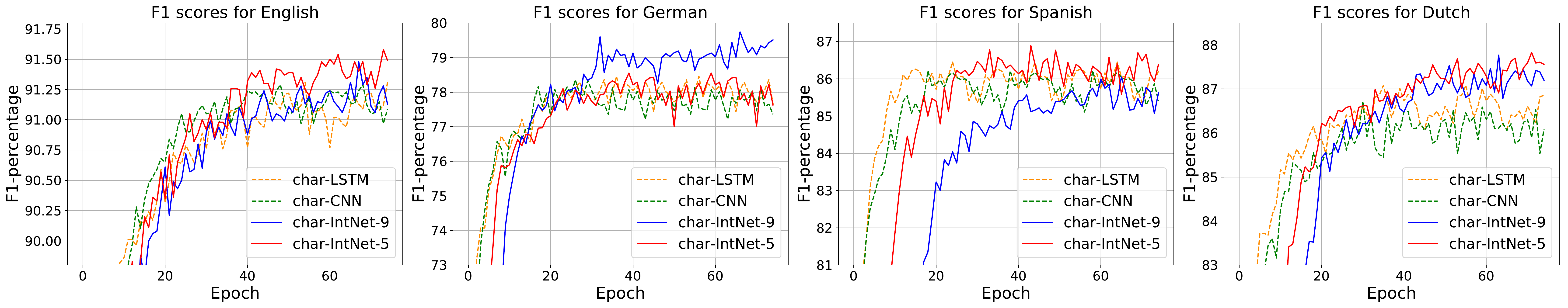}
  \caption{Training details of different models for English, German, Spanish, and Dutch.}
  \label{fig-f1-training}
\end{figure*}

\subsection{Compared Methods} 

To address those open questions in Section \ref{sec:intro}, we conduct detailed experiments and empirical comparisons on different state-of-the-art character embedding models across different domains. Firstly, we use LSTM-CRF with randomly initialized word embeddings as our initial baseline. We adopt two state-of-the-art methods in sequence labeling, denoted as char-LSTM \cite{lample2016neural} and char-CNN \cite{ma2016end}. We add more layers to the char-CNN model and refer to that as char-CNN-5 and char-CNN-9, respectively for 5 and 9 convolutional layers. Furthermore, we add residual connections to the char-CNN-9 and refer it as char-ResNet. Also, we apply 3 dense blocks based on char-ResNet which we refer to as char-DenseNet, to compare the difference between residual connection and dense connection. Lastly, we refer to our proposed model, which uses different convolution layers, as char-IntNet-5 and char-IntNet-9.

\section{Results and Analysis}
\label{sec:results}

\subsection{Character-to-word Models}

Table \ref{tab-compared-models} presents the performance of different character-to-word models on six benchmark datasets. For sequence labeling, char-LSTM and char-CNN are current state-of-the-art character embedding models for learning character-to-word representations. We observe that char-LSTM performs better than char-CNN in most cases, however, char-CNN uses a convolution layer followed by max pooling as a shallow feature extractor, that does not explore the full potential of CNNs. 

Therefore, we implement two variations based on char-CNN, referred to as char-CNN-5 and char-CNN-9. The result shows that for most of the datasets, the F1 score does not improve much when we directly add more layers. We also observe some accuracy drop when we continuously increase the depth. This confirms why most CNN-based approaches for learning representations on words are shallow, which is very different from learning representations for sentences. Furthermore, we add residual connections to char-CNN-9 as char-ResNet-9, which confirms that residual connections can help train deep layers. We further improve char-ResNet-9 by changing residual connections into dense connection blocks as char-DenseNet-9, which shows that the dense connections are better than residual connections for learning word shape information.

Our proposed character-to-word model, char-IntNet-5 and char-IntNet-9 generally improves the results across all datasets. Our IntNet significantly outperforms other character embedding models, for example, the improvement is more than 2\% in terms of F1 score for German and Dutch. Also, we observe that char-IntNet-5 is more effective for learning character-to-word representations than char-IntNet-9 in most of the cases. The only exception is German which seems to require a deeper and wider model for learning better representations.

\begin{table*}[htb]
\centering
\scalebox{0.59}{
\begin{tabular}{|c|c|c|c|c|c|c|c|c|c|c|c|c|c|c|c|c|c|c|c|c|}
\hline
 &\multirow{2}{*}{Model} &\multicolumn{4}{c|}{English}  &\multicolumn{4}{c|}{German}  &\multicolumn{4}{c|}{Spanish}  &\multicolumn{4}{c|}{Dutch} \\\cline{3-18}

 & &IV &OOTV &OOEV &OOBV    &IV &OOTV &OOEV &OOBV    &IV &OOTV &OOEV &OOBV    &IV &OOTV &OOEV &OOBV \\ \hline

\multirow{4}{*}{Dev} 
&char-LSTM     &97.15 &89.87 &89.41 &87.07   &86.97 &85.80 &68.35 &64.76   &89.63 &89.06 &78.14 &74.13   &94.50 &87.98 &80.00 &72.37 \\
&char-CNN      &97.10 &90.04 &95.45 &88.02   &87.45 &86.13 &57.14 &63.28   &88.93 &88.85 &72.90 &71.96   &94.54 &87.27 &74.55 &68.77 \\
&char-IntNet-9 &96.86 &\textbf{90.52} &\textbf{91.95} &\textbf{90.16}   &\textbf{87.92} &85.29 &\textbf{76.07} &\textbf{67.98}   &88.43 &\textbf{88.58} &74.53 &\textbf{72.09}   &93.68 &87.49 &\textbf{89.09} &\textbf{75.58} \\
&char-IntNet-5 &96.65 &\textbf{90.14} &88.10 &\textbf{88.31}   &87.21 &85.00 &67.10 &64.17  &88.56 &88.47 &\textbf{78.90} &70.23   &\textbf{94.63} &\textbf{88.56} &\textbf{89.09} &\textbf{74.40} \\ \hline

\multirow{4}{*}{Test} 
&char-LSTM     &93.68 &92.48 &100.00 &82.64   &86.97 &83.95 &69.67 &62.74   &87.19 &87.79 &95.29 &76.01   &95.13 &83.00 &78.26 &72.34 \\
&char-CNN      &93.85 &92.65 &100.00 &84.09   &64.72 &83.67 &69.67 &58.19   &87.81 &88.46 &87.96 &73.68   &94.25 &82.50 &73.27 &73.37 \\
&char-IntNet-9 &93.79 &\textbf{94.94} &100.00 &82.31   &\textbf{87.56} &\textbf{83.85} &\textbf{74.33} &\textbf{65.75}   &87.08 &87.98 &\textbf{95.29} &\textbf{77.16}   &94.42 &\textbf{83.85} &\textbf{85.02} &\textbf{75.46} \\

&char-IntNet-5 &\textbf{93.94} &\textbf{92.72} &100.00 &\textbf{83.91}   &\textbf{87.11} &83.60 &67.22 &60.92   &87.19 &\textbf{88.42} &\textbf{97.38} &\textbf{78.02}   &94.71 &\textbf{84.84} &\textbf{82.13} &\textbf{76.99} \\ \hline

\end{tabular}
}
\caption{F1 score of different models for IV, OOTV, OOEV and OOBV.}
\label{tab-oov}
\end{table*}

\begin{table*}[htb]
\centering
\scalebox{0.675}{
\begin{tabular}{|c|c|c|c|c|c|c|c|c|c|}
\hline
\multirow{2}{*}{Model}  &\multicolumn{3}{c|}{Frequent Words}  &\multicolumn{3}{c|}{Rare Words}  &\multicolumn{3}{c|}{OOV Words} \\\cline{2-10}
\multirow{7}{*}{char-LSTM}
&$newspapers$ &$slipped$ &$world$ &$Commerce$ &$youthful$ &$sessions$ &$11$-$month$ &$Thursdays$ &$undetermined$ \\ \hline
&enclosures &stirred &wolrd &Committee &luthier &cessions &19-month &Thousands &undereducated \\ 
&nelsonville &clipped &worde &Computer &loughmoe &sensible &10-month &Tunbridge &underpinned \\ 
&entrances &snipped &lowed &Comments &wrathful &stepanos &12-month &Standings &undermined \\
&newpapers &striped &wowed &Corrects &slothful &stefanos &14-month &Torrance &underlined \\
&necklaces &stifled &crowd &Clippers &ephorus &constans &11-inch &Phillies &underprepared \\ \hline
\multirow{5}{*}{char-CNN}
&newspaper &slipper &worli &Committee &mouthful &suppressions &31-month &Thursday &determined \\ 
&newspapermen &slippy &worle &Community &eeyou &oppressions &51-month &Wednesday &overdetermined \\ 
&newpapers &stripped &worse &Commodities &mouthfeel &digressions &1-month &Tuesday &determinist \\ 
&nitrification &shipped &werle &Communist &motul &confessions &21-month &Ecuador &determiners \\
&megaphones &stopped &wereld &Comments &yourself &fissions &41-month &Windass &determiner \\ \hline
\multirow{5}{*}{char-IntNet}
&newpapers &blipped &eworld &Commissioner &mouthful &recessions &55-month &Thursday &undermined \\ 
&wallpapers &unclipped &offworld &Commodities &mirthful &accessions &51-month &Saturday &determined \\ 
&escapers &tripped &homeworld &Clarence &mouthfuls &missions &22-month &thursdays &overdetermined \\ 
&carcases &dripped &linuxworld &Commission &youths &conversions &25-month &Tuesday &unexamined \\
&spacers &slopped &westworld &Commons &slothful &possessions &12-month &tuesdays &predetermined \\ \hline
\end{tabular}
}
\caption{Nearest neighbours of different models for frequent words, rare words and OOV words.}
\label{tab-word-knn}
\end{table*}

\subsection{State-of-the-art Results}

Table \ref{tab-state-of-the-art} presents our proposed model in comparison with state-of-the-art results. LSTM-CRF is our baseline which uses fine-tuned pre-trained word embeddings. Its comparison with LSTM-CRF using random initializations for word embeddings, as shown in Table \ref{tab-compared-models}, confirms that pre-trained word embeddings are useful for sequence labeling. Since the training corpus for sequence labeling is relatively small, pre-trained embeddings learned from a huge unlabeled corpus can help to enhance word semantics. Furthermore, we adopt and re-implement two state-of-the-art character models, char-LSTM and char-CNN, by combining with LSTM-CRF, which we refer to as LSTM-CRF-char-LSTM and LSTM-CRF-char-CNN. Lastly, we combine our proposed model with LSTM-CRF which we refer to as LSTM-CRF-char-IntNet-9 and LSTM-CRF-char-IntNet-5. 

These experiments show that our char-IntNet generally improves results across different models and datasets. The improvement is more pronounced for non-English datasets, for example, IntNet improves the F-1 score over the state-of-the-art results by more than 2\% for Dutch and Spanish. It also shows that the results of LSTM-CRF are significantly improved after adding character-to-word models, which confirms that word shape information is very important for sequence labeling. Figure \ref{fig-f1-training} presents the details of training epochs in comparison with other state-of-the-art character models for different languages. It shows that char-CNN and char-LSTM converge early whereas char-IntNet takes more epochs to converge and generally performs better. It alludes to the fact that IntNet is suitable for reducing over-fitting,  since we have used early stopping while training.

\subsection{Rare and OOV Words Analysis}
\label{sec:oov} 

Another advantage of learning internal structure of words is that it can capture representations for out-of-vocabulary (OOV) words. To better understand the behavior of IntNet, Table \ref{tab-oov} presents error analysis on in-vocabulary words (IV), out-of-training-vocabulary words (OOTV), out-of-embedding-vocabulary words (OOEV), and out-of-both-vocabulary words (OOBV) compared to different character models. The result shows that our proposed model significantly outperforms other character models on OOV words including OOTV, OOEV, and OOBV. For example, in OOBV category, our IntNet outperforms other models by more than 3\% in terms of F1 score for Dutch and German datasets.

Furthermore, we present comparisons of nearest neighbors with different models for frequent words, rare words, and OOV words. Table \ref{tab-word-knn} shows the results of nearest neighbors for learning word shape information, which gives insights on what kind of character-to-word representations can be learned by different models. For example, in OOV words, our IntNet model learns a better $xx$-$month$ shape pattern when matching $11$-$month$ compared to other models.

\subsection{Discussion}

In many situations, learning character-to-word representations of subword sequences that exceed the typical length of word shape pattern or morpheme sequences might result in noise. RNNs can capture longer sequences in theory, however, longer sequences do not guarantee better results when learning prefixes and suffixes. The funnel-shaped wide architecture of IntNet, uses different kernels with different levels of concatenation to capture patterns of different subword lengths and that is flexible than char-LSTM and char-CNN. For example, Table \ref{tab-word-knn} shows $Thursday$ in OOV words, our model learns a better word-shape structure for character-to-word representations compared to other methods.

When considering training time, IntNet is only 20\% slower than char-CNN for the whole training process. Also, learning word representations use fewer parameters than learning sentence representations. Therefore, the impact of training speed for sequence labeling is limited. The inference time of IntNet is almost the same as char-CNN.

\section{Conclusion}

We presented empirical comparisons of different character embedding models for learning character-to-word representations and investigated the gaps between methods for learning representations of words and sentences. We conducted detailed experiments of different state-of-the-art convolutional models, and explored the advantages and disadvantages of their components for learning word shape information. Furthermore, we presented IntNet, a funnel-shaped wide convolutional neural architecture with no down-sampling that focuses on learning better internal structure of words by composing their characters from limited supervised training corpora. Our in-depth analysis showed that a shallow wide architecture is better than a narrow deep architecture for learning character-to-word representations. Omitting down-sampling operations is useful for capturing the sequential patterns of prefixes and suffixes. Our proposed compositional character-to-word model does not leverage any external resources, hand-crafted features, additional knowledge, joint training, or character-level preprocessing, and achieves new state-of-the-art performance for various sequence labeling tasks, including named entity recognition, part-of-speech tagging and syntactic chunking. In the future, we would like to explore using the IntNet model for other NLP tasks.

\bibliography{emnlp2018}
\bibliographystyle{acl_natbib_nourl}

\end{document}